\documentclass[onecolumn, a4paper,journal,12pt]{IEEEtran}
\usepackage{amsfonts}
\usepackage{graphicx, epsfig,amsmath,amssymb,latexsym,graphics,float}
\usepackage{setspace,subfig}
\usepackage{citesort}

\include{IEEEtran.bib}

\begin{document}

\title{A Very Brief and Critical Discussion on AutoML}
\author{Bin~Liu\\
School of Computer Science\\Nanjing University of Posts and Telecommunications\\Nanjing, 210023 China\\
Email: bins@ieee.org\\
}
\maketitle
\begin{abstract}
This contribution presents a very brief and critical discussion on automated machine learning (AutoML), which is categorized here into two classes, referred to as narrow AutoML and generalized AutoML, respectively. The conclusions yielded from this discussion can be summarized as follows: (1) most existent research on AutoML belongs to the class of narrow AutoML; (2) advances in narrow AutoML are mainly motivated by commercial needs, while any possible benefit obtained is definitely at a cost of increase in computing burdens; (3)the concept of generalized AutoML has a strong tie in spirit with artificial general intelligence (AGI), also called ``strong AI", for which obstacles abound for obtaining pivotal progresses. 
\end{abstract}

\section{Introduction}\label{sec:intro}
AutoML has recently emerged as a hot research topic in the field of machine learning (ML) and artificial intelligence (AI). As we know, a typical ML pipeline requires a lot of human's participation for e.g., data pre-processing, feature engineering, algorithm selection, model selection and hyperparameter optimization. The purpose of AutoML is to make the ML pipeline automated, getting rid of the aforementioned cumbersome issues that are often beyond the abilities of non-experts.

The notion AutoML was first introduced in an ICML-2014 workshop \cite{Hutter2014}, while the idea of AutoML appeared earlier in e.g. \cite{thornton2013auto}. Some classical examples of AutoML implementations include the Bayesian optimization (BO) based automation of WEKA (Auto-WEKA) \cite{thornton2013auto} and sklearn (auto-sklearn) \cite{feurer2015efficient}, genetic programming based automation of ML pipelines using Python library TPOT \cite{olson2016automating} and the RECIPE framework \cite{de2017recipe}, and Google's Cloud AutoML, which can do adaptive neural architecture search \cite{li2018cloud}.

Although intuition tells us that the idea of AutoML is cool and it has indeed gained some successes in some applications, I shall provide several critical discussions on it, hoping to stimulate more reasonable considerations and discussions on the concept of and technologies about AutoML.
\section{Discussions}
\subsection{A Classification of AutoML}
In the first AutoML workshop, AutoML was described a research area that targets progressive automation of ML \cite{Hutter2014}. The term ``progressive" indicates that it is likely a long-term goal to implement fully automated ML. Here I categorize all possible AutoML techniques into two classes, referred to as narrow AutoML and generalized AutoML, respectively. The former denotes all intermediate ML techniques developed under the way in achieving fully automated ML and the latter represents the fully automated ML. In contrast with traditional ML, narrow AutoML can reduce the amount of but cannot fully get rid of experts' involvement, while if generalized AutoML is ideally implemented, then no expert effort is needed to perform an ML task.
\subsection{On narrow AutoML}
I argue that most existent progresses in the area of AutoML are just progresses in the sub-field of narrow AutoML. Take the BO based automatic algorithm configuration as an example for analysis. BO has been successfully used for hyperparameter optimization \cite{shahriari2016taking}, model selection \cite{malkomes2016bayesian}, structure tuning of neural networks \cite{mendoza2016towards} and so on. The basic idea of BO is to substitute the expert's effort with an outer-loop algorithm, e.g., Gaussian Process (GP) regression, for configuring hyperparameters or model structures for the inner-loop ML algorithm of interest. Hence the outer-loop GP based searching procedure can be regarded as a surrogate of the expert, whose task is to seek appropriate hyperparameter values or model structures for the inner-loop algorithm. The BO based AutoML is just a kind of narrow AutoML because as an outer-loop algorithm, the BO itself has hyperparameters, e.g., the kernel type, the acquisition function, that need to be configured by an expert or an outer-loop algorithm which acts as a surrogate of the expert. For example, in \cite{feurer2015initializing}, a meta-learning technique is adopted for initializing BO, while both of the selection and the parameter configuration of the meta-learning technique are done by human expert. In \cite{gardner2017discovering,malkomes2016bayesian}, a compositional kernel searching strategy is developed to automate BO, while it still requires expert knowledge to specify the space of compositional kernels as well as hyper parameter priors. 

Although the above analysis is only restricted to BO, the same result holds for other types of automatic algorithm configuration techniques, e.g., transfer learning, reinforcement learning, heuristic methods, which are mentioned in \cite{quanming2018taking}.
To summarize, for most existent AutoML works, regardless of the number of layers of the outer-loop algorithms, the configuration of the outermost layer is definitely done by human experts. So they all belong to the narrow AutoML class. It is shown that employing such narrow AutoML methods, the amount of required expert knowledge can be reduced but cannot be avoided.

Another point to note is that any benefit gained by reducing expert's involvements is at the cost of the increase in the computing burdens. Still taking the BO based AutoML as an instance, we can see that any automation brought by BO is implemented through expanding the searching space of hyperparameters or models or both for the inner-loop ML algorithm. Consider an ideal extreme case in which we have a qualified expert who can determine the values of all hyperparameters and the model structure very quickly totally based on his experience, then no additional computing burden except running the algorithm is needed to finish the learning task. Then if we now substitute this expert with an out-loop BO algorithm, then we have to pay corresponding computing burdens to initializing the BO and letting it search the value space of the hyperparameters and model structures. 
\subsection{On generalized AutoML}
As aforementioned, narrow AutoML is characterized by algorithmically configuration of an inner-loop ML algorithm through invoking an outer-loop algorithm, and thus can not achieve fully automated ML, since the configuration of the outermost layer algorithm has to be configured by human experts. In contrast, the concept of generalized AutoML means fully automated ML without expert's involvement or expert knowledge, which represents the final goal of the AutoML research, as stated in recent AutoML workshops at ICML \cite{Hutter2014,Garnett2018,Adams2016}. I argue that the concept of generalized AutoML has a strong tie in spirit with AGI, also called ``strong AI", for which obstacles abound for obtaining pivotal progresses. Details on AGI are referred to \cite{goertzel2007artificial,goertzel2007foundational,wang2007introduction,voss2007essentials}.
\section{Conclusions}
In this brief note, I categorized the concept of AutoML into two classes, namely narrow AutoML and generalized AutoML, and showed that most existent advances in the area of AutoML are just within the scope of narrow AutoML. I also provided a cautious reminder of that the employment of narrow AutoML also requires an amount of expert's involvement or expert knowledge, and any benefit obtained is definitely at a cost in the increase of computing burdens. Finally, I pointed out that the concept of generalized AutoML is closely related with AGI, for which obstacles abound for obtaining pivotal progresses.
\bibliographystyle{IEEEtran}
\bibliography{mybibfile}
\end{document}